\documentclass[letterpaper]{article}
\usepackage[preprint]{aaai2027}   % preprint: no copyright slug, authors shown
\usepackage[hyphens]{url}
\usepackage{graphicx}
\usepackage{natbib}
\usepackage{caption}
\usepackage{newfloat}
\usepackage{listings}
\DeclareCaptionStyle{ruled}{labelfont=normalfont,labelsep=colon,strut=off}
\usepackage{booktabs}
\usepackage{multirow}
\usepackage{amsmath, amssymb}
\usepackage{xcolor}
\usepackage{xspace}

\usepackage{fancyhdr}

\newcommand{\reze}{\textsc{REZE}\xspace}

\newcommand{\miou}{mIoU\xspace}

\title{REZE: Recognition-Based Zero-Shot Extraction for Video Temporal Grounding}

\author{
    Boyang Li,
    Chenhui Gou,
    Jianfei Cai\corresponding
}

\affiliations{
    Department of Data Science and AI, Monash University\\
    blii0116@student.monash.edu,
    \{chenhui.gou, jianfei.cai\}@monash.edu
}

\begin{document}

\maketitle

\begin{abstract}
Video temporal grounding (VTG) refers to the task of identifying the time interval in a video that corresponds to a given natural-language query.
A common zero-shot strategy asks a large vision-language model (VLM) to generate the start and end timestamps directly, so the result depends heavily on the design and training of the model, and grounding accuracy differs widely from one VLM to another.
We therefore propose REcognition-based Zero-shot Extraction (REZE), a simple training-free method that splits the video into short clips, asks the model for a clip-level confidence score for the query, and uses a deterministic algorithm to convert the resulting score curve into the output required by the task.
Because temporal aggregation is performed outside the model, REZE adapts to different task outputs, from single- and multi-interval moment retrieval to highlight detection.
On QVHighlights, REZE improves the best reported training-free moment-retrieval mAP from $38.23$ to $40.32$, while on highlight detection it reaches $44.18$ mAP and $73.41$ HIT@1, establishing a new state of the art among training-free methods.
Its HIT@1 also outperforms all fully supervised SoTAs on the QVHighlights test split.
We evaluate REZE on seven backbones from three model families.
On Charades-STA and QVHighlights, it outperforms direct timestamp generation in every available comparison.
We further observe that with REZE an earlier-generation model can approach the native performance of a newer model in its family.
\end{abstract}

\section{Introduction}
\label{sec:intro}

\begin{figure*}[t]
\centering
\includegraphics[width=0.93\textwidth]{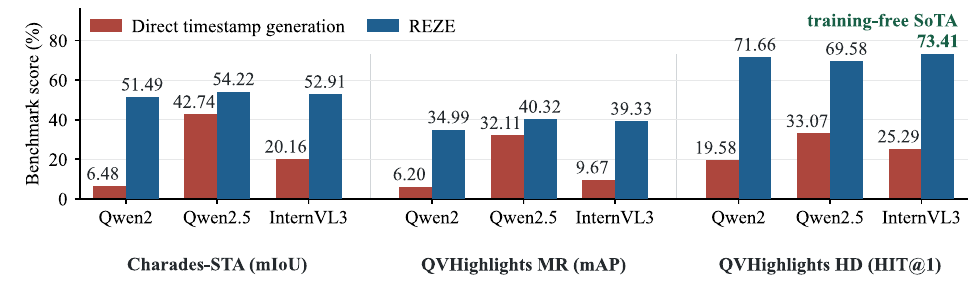}
\caption{Direct timestamp generation (\emph{Direct}) versus \reze{} across three VTG outputs and
three backbones~\citep{wang2024qwen2vl,bai2025qwen25vl,chen2025internvl3}.
Each adjacent bar pair compares the same backbone, with Direct
on the left and \reze{} on the right. The blocks report Charades-STA \miou{},
QVHighlights MR mAP, and QVHighlights HD HIT@1. Direct generation has no native
highlight output, so its predicted intervals are converted into a per-clip
saliency mask. Exact scores are printed above the bars, and the marker identifies
the training-free state of the art in HIT@1. Results use the benchmark test
splits.}
\label{fig:headline}
\end{figure*}

Video temporal grounding (VTG) identifies when the content described by a
natural-language query occurs in a video. Depending on the benchmark, the
required output may be a single interval, several relevant intervals, or a
saliency score for each video clip~\citep{gao2017tall,lei2021qvhighlights}.
Despite these different output formats, every setting requires the system to
measure how strongly the query matches the video over time. The system must
then convert this evidence into the output expected by the task.

Existing methods handle these requirements in two main ways. Trained grounding
models learn to localise queries from annotated
videos~\citep{zhang20202dtan,zeng2020drn,mun2020lgi,li2025unitime}. Training-free
approaches instead reuse a frozen model and obtain temporal predictions through
prompting, visual overlays, or clip-level similarity and confidence
scores~\citep{zheng2024tfvtg,wu2025numpro,tag2025,aklilu2024zeal}. Among these
approaches, a common strategy is direct timestamp generation (\emph{Direct}), in which a VLM
returns the start and end timestamps as text~\citep{wang2024qwen2vl,bai2025qwen25vl,qwen3vl}.
This strategy avoids task-specific training; however, the model still has to
recognise the queried content and generate its timestamps in the same response,
and thus an error in either step may lead to an incorrect prediction.

Instead of asking a VLM to recognise the queried content and generate its
timestamps in the same response, we propose REcognition-based Zero-shot
Extraction (REZE), a simple training-free method that separates these two
steps. REZE uses a frozen VLM to evaluate the query on short clips and obtains
a confidence score from each clip. When ordered along the video timeline,
these scores describe how the model's confidence in the queried content changes
over time. A deterministic temporal aggregation algorithm then converts the
resulting score sequence into the output required by the task.

Because temporal aggregation is performed outside the VLM, REZE can adapt the
same score sequence to different task outputs by changing only the final
aggregation rule. For single-interval moment retrieval, the aggregation extracts
one interval, whereas for multi-interval moment retrieval it can return multiple
disjoint intervals. Highlight detection requires no interval extraction and
instead uses the same scores to rank clips by saliency. The video segmentation,
recognition question, and score extraction therefore remain unchanged across
all three settings.

We evaluate REZE on Charades-STA, ActivityNet Captions, and
QVHighlights~\citep{gao2017tall,krishna2017densecaptioning,lei2021qvhighlights},
covering single-interval and multi-interval moment retrieval as well as
highlight detection. Figure~\ref{fig:headline} compares REZE with Direct timestamp
generation across the three output formats. On QVHighlights, REZE establishes
new training-free state-of-the-art results on both tasks, reaching $40.32$ mAP
for moment retrieval and $44.18$ mAP and $73.41$ HIT@1 for highlight
detection~\citep{granalign,wu2025numpro}. Its HIT@1 outperforms all fully
supervised SoTAs on the QVHighlights test
split~\citep{hlclip,cva,sgdetr}. On ActivityNet Captions, REZE raises
Qwen2.5-VL from $12.61$ to $36.47$ \miou{}, approaching the $36.76$ achieved by
Qwen3-VL under Direct~\citep{qwen3vl}. On Charades-STA, REZE outperforms Direct
on all six VLMs that support native timestamp generation. REZE also enables the
image-only LLaVA-1.5~\citep{liu2024llava} to perform VTG even though the model
has no native Direct setting.

\paragraph{Contributions.}
Our contributions are threefold.
(i)~We propose REZE, which separates clip-level recognition from temporal
aggregation and applies to both video-capable and image-only VLMs without
task-specific training.
(ii)~We evaluate REZE across seven VLMs from three families and show that it
improves every available Direct comparison on Charades-STA and QVHighlights;
its HIT@1 on QVHighlights highlight detection sets a new state of the art among
training-free methods.
(iii)~We demonstrate that the same method supports single-interval and
multi-interval moment retrieval as well as highlight detection by changing only
the final aggregation rule.

\section{Related Work}
\label{sec:background}

\paragraph{Supervised video temporal grounding.}
Supervised VTG learns temporal predictions from videos and queries paired with interval or saliency annotations. Early work established single-interval retrieval on Charades-STA and developed stronger ways to match temporal candidates with a query~\citep{gao2017tall,zhang20202dtan,zeng2020drn,mun2020lgi}. QVHighlights later introduced multi-interval moment retrieval and highlight detection in one benchmark~\citep{lei2021qvhighlights}. On this benchmark, QD-DETR, CG-DETR, and UVCOM improve query--clip matching, while R2-Tuning and SDST adapt pretrained visual encoders~\citep{qd_detr,cg_detr,uvcom,r2tuning,sdst}. Our highlight-detection comparisons further include HL-CLIP, CVA, SG-DETR, and CoSTL, which also use QVHighlights annotations during training~\citep{hlclip,cva,sgdetr,costl}. Video VLMs such as TimeChat, VTimeLLM, TRACE, Time-R1, and UniTime likewise learn grounding from timestamp-annotated data~\citep{ren2024timechat,huang2023vtimellm,guo2024trace,timer1,li2025unitime}. In particular, UniTime uses ``zero-shot'' to describe transfer to a new dataset without target-specific fine-tuning, but the model has already been trained for temporal grounding. MeCo avoids timestamp generation by training a video VLM to generate structural tokens and ground them to event segments~\citep{pang2026measuretwicecutonce}. REZE instead performs temporal grounding without task-specific training.

\paragraph{Training-free video temporal grounding.}
Other training-free methods also keep pretrained models frozen. TFVTG decomposes a query into ordered sub-events, while NumPro overlays numeric indices on video frames~\citep{zheng2024tfvtg,wu2025numpro}. TAG, Robust Zero-Shot, and GranAlign construct intervals from video--text similarity or alignment scores~\citep{tag2025,robustzs,granalign}. The Robust Zero-Shot result used in our Charades-STA comparison is a three-model ensemble, whereas GranAlign provides the strongest prior training-free QVHighlights moment-retrieval result in our comparison. ZEAL is closest to REZE because both convert VLM confidence scores into temporal predictions~\citep{aklilu2024zeal}. However, ZEAL expands an action category into descriptions of its typical start and end states and scores sampled frames against them. REZE instead scores each short clip against the original natural-language query and aggregates the ordered scores into a single interval, multiple intervals, or clip-level saliency.

\paragraph{Video and timestamp interfaces in VLMs.}
The evaluated backbones expose different native interfaces. The Qwen-VL series accepts video and generates textual timestamps, the InternVL series treats sampled frames as multi-image input, and image-only LLaVA-1.5 has no native timestamp output~\citep{wang2024qwen2vl,bai2025qwen25vl,qwen3vl,chen2025internvl25,chen2025internvl3,wang2025internvl35,liu2024llava}. REZE applies the same clip-level scoring and external aggregation to all three interfaces. It therefore does not ask the VLM to order evidence across the full video, a capability that prior work has found challenging for several tested video-language models~\citep{bagad2023testoftime,jung2025consistency}.

\section{Method: REZE}
\label{sec:method}

\subsection{Problem Formulation}

Given an untrimmed video $V$ of duration $D$ and a natural-language query $q$, video temporal grounding (VTG) identifies when the queried content occurs. Although this objective is shared across benchmarks, the required output varies by task. Charades-STA~\citep{gao2017tall} and ActivityNet Captions~\citep{krishna2017densecaptioning} require one interval $(\hat{t}_s,\hat{t}_e)$, where $0 \leq \hat{t}_s < \hat{t}_e \leq D$. By contrast, QVHighlights moment retrieval~\citep{lei2021qvhighlights} requires a ranked list because one query can correspond to multiple relevant moments. In addition, QVHighlights highlight detection assigns a saliency score to each clip.

\begin{figure*}[t]
\centering
\includegraphics[width=\textwidth]{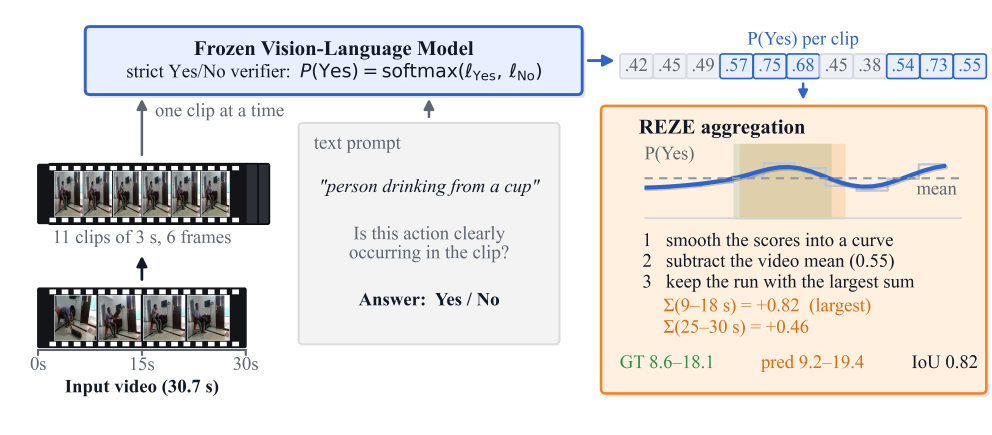}
\caption{The \reze{} pipeline, illustrated with a single-interval example from Charades-STA. The $30.7$-second video is divided into $11$ three-second clips, each represented by six frames. A frozen VLM evaluates every clip with the same binary prompt and produces one $P(\text{Yes})$ score, yielding the $11$ temporally ordered scores shown on the right. Deterministic aggregation outside the VLM then converts these scores into the predicted interval. Multi-interval moment retrieval and highlight detection use the same score sequence with different external readouts.}
\label{fig:reze-pipeline}
\end{figure*}

To separate content recognition from task-specific output construction, REZE supports all three outputs through the same intermediate representation. It divides $V$ into $N$ temporally ordered clips $\{F_i\}_{i=1}^{N}$, where $N$ is the number of clips and $F_i$ is the $i$-th clip. A frozen VLM assigns each $F_i$ a confidence score $p_i\in[0,1]$ for the query $q$. These scores form the ordered sequence $P=(p_1,\ldots,p_N)$, which describes how the model's confidence changes over the video. External temporal aggregation then maps $P$ to the output required by the task. Figure~\ref{fig:reze-pipeline} summarises the pipeline.

\subsection{Clip-Level Confidence Scoring}
\label{sec:method-pipeline}

\paragraph{Clip construction.}
REZE divides $V$ into consecutive, non-overlapping clips with maximum duration $\tau=3$ seconds and samples each clip at $f=2$ frames per second. Thus, a complete clip contains six evenly spaced frames. We use the same $\tau$ and $f$ for every backbone and task. Appendix~A gives the exact segmentation procedure, while Appendix~C shows that reducing $\tau$ to $2$ seconds improves mAP by only about $0.7$ points but requires $1.5\times$ as many clip evaluations.

\paragraph{Binary verification prompt.}
For each clip $F_i$, REZE asks whether the queried content is clearly and actively present. The asymmetric prompt permits \texttt{Yes} only when this condition is satisfied and uses \texttt{No} as the default response, which is intended to suppress high scores on background clips. We use the same prompt for all backbones and tasks. Appendix~A provides the complete prompt, and Appendix~C reports a wording sweep.

\paragraph{Continuous clip score.}
Instead of turning the model's answer into a hard decision, REZE reads the next-token logits $\ell_{\text{Yes}}^{(i)}$ and $\ell_{\text{No}}^{(i)}$ and normalises them with a two-way softmax:
\begin{equation}
p_i
=
P(\text{Yes}\mid F_i,q)
=
\frac{\exp(\ell_{\text{Yes}}^{(i)})}
{\exp(\ell_{\text{Yes}}^{(i)})+\exp(\ell_{\text{No}}^{(i)})}.
\label{eq:pyes}
\end{equation}
The resulting value $p_i$ is used as a relative confidence score for temporal aggregation, not as a calibrated probability over the full vocabulary. Reading the logits directly also avoids asking the VLM to generate a numerical score. Moreover, keeping $p_i$ continuous preserves confidence differences that hard thresholding would remove. Appendix~A gives the exact tokenisation and extraction procedure, while Appendix~C compares continuous and binary scores.

\paragraph{Backbone interfaces.}
A multi-frame backbone receives all sampled frames from $F_i$ together and produces one score $p_i$. By contrast, an image-only backbone scores the frames separately, after which REZE averages them to obtain the same clip-level output. The distinction affects only how $p_i$ is obtained; the downstream aggregation is unchanged. Both interfaces therefore produce the ordered sequence $P$ without additional training or a cross-clip temporal model.

\subsection{External Temporal Aggregation}
\label{sec:method-readout}

\paragraph{Per-second score curve.}
To aggregate the independently obtained clip scores, REZE first converts $P$ into a per-second curve $s(t)$. Here, $t\in\{0,\ldots,\lfloor D\rfloor-1\}$ indexes integer seconds, and $s(t)\in[0,1]$ is the confidence assigned to second $t$. Each second receives the score of the clip that covers it. The resulting curve restores the temporal order of the clip-level evidence and serves as the common input to three output-specific readouts. All subsequent operations are deterministic and performed outside the VLM. Appendix~A specifies the boundary and interpolation rules used to construct $s(t)$.

\paragraph{Single-interval output.}
For a task that requires one interval, we smooth $s(t)$ with a one-dimensional Gaussian kernel. Its standard deviation is $\sigma=\max(1,\lfloor D\rfloor/15)$ seconds, and we denote the smoothed curve by $\bar{s}(t)$. We use this duration-dependent rule for both single-interval benchmarks rather than tuning a separate bandwidth for each dataset. Across the reported test backbones, fixed $2$- or $3$-second kernels change \miou{} by at most $1.3$ points on Charades-STA and $1.5$ points on ActivityNet Captions. By contrast, removing smoothing entirely can reduce \miou{} by $4.2$ points (Appendix~C). We then subtract the video mean:
\begin{equation}
s'(t)
=
\bar{s}(t)
-
\frac{1}{\lfloor D\rfloor}
\sum_{u=0}^{\lfloor D\rfloor-1}\bar{s}(u),
\label{eq:meanshift}
\end{equation}
where $u$ indexes all integer seconds and $s'(t)$ is the mean-centred score. Mean-centring prevents the non-negative curve from favouring the entire video. It also makes positive values represent above-average evidence and negative values penalise below-average regions. REZE then uses Kadane's algorithm~\citep{bentley1984} to find the half-open range $[t_s^\star,t_e^\star)$ that maximises $\sum_{t=t_s^\star}^{t_e^\star-1}s'(t)$. This objective accumulates evidence over time, allowing a sustained region of moderate confidence to outweigh an isolated peak. Finally, the selected range is mapped back to the original video timeline to obtain $(\hat{t}_s,\hat{t}_e)$. Section~\ref{sec:exp-analysis} compares this interval extractor with nine alternatives.

\paragraph{Multi-interval output.}
QVHighlights moment retrieval can require several disjoint intervals. For this task, a duration-dependent kernel can become broad on long videos and blur shorter relevant regions. The multi-interval readout is also more sensitive to this scale than the single-interval readout. With the threshold fixed, shortening $\sigma$ from $\lfloor D\rfloor/15$ to $3$ seconds raises QVHighlights validation mAP by $16.0$ points (Appendix~C). We therefore smooth $s(t)$ at the clip scale using $\sigma=\tau=3$ seconds, so the smoothing width does not grow with video duration. Otsu's method~\citep{otsu1979} then determines a threshold $\theta$ from the values of the smoothed curve $\bar{s}(t)$. Because $\theta$ is estimated separately for each curve, the readout adapts to its score distribution without using a fixed global threshold. Contiguous regions above $\theta$ become ranked candidate intervals, which allows one query to produce several disjoint outputs. This readout is fixed across backbones and test queries. Appendix~A gives the filtering, ranking, and fallback rules, while Appendix~C reports the smoothing and threshold ablations.

\paragraph{Highlight detection.}
Highlight detection uses the unsmoothed curve $s(t)$ because it requires no interval boundaries. Since QVHighlights evaluates consecutive two-second clips~\citep{lei2021qvhighlights}, REZE averages $s(t)$ over each evaluation clip and uses the result as its saliency score. Thus, the same sequence $P$ supports one interval, multiple intervals, and clip-level saliency by changing only the deterministic external readout.

\section{Experiments}
\label{sec:experiments}

\subsection{Experimental Setup}

\paragraph{Benchmarks and metrics.}
We evaluate REZE on three VTG benchmarks that cover all three output formats
considered in this work. Charades-STA~\citep{gao2017tall,sigurdsson2016charades}
($3{,}720$ test queries) and ActivityNet Captions~\citep{krishna2017densecaptioning}
(val\_2, $17{,}031$ queries) evaluate single-interval retrieval, for which we
report mean intersection over union (\miou{}) and Recall at IoU thresholds of
$0.3$, $0.5$, and $0.7$. QVHighlights~\citep{lei2021qvhighlights} evaluates
multi-interval moment retrieval and clip-level highlight detection on its test
split ($1{,}542$ queries). Following the benchmark protocol, we report
moment-retrieval mAP, and for highlight detection, we report mAP and HIT@1.

\paragraph{Models and comparisons.}
Across these benchmarks, we evaluate seven pretrained VLMs from three model
families: the image-only LLaVA-1.5-7B~\citep{liu2024llava}, Qwen2-VL-7B,
Qwen2.5-VL-7B, and Qwen3-VL-8B~\citep{wang2024qwen2vl,bai2025qwen25vl,qwen3vl},
and InternVL2.5-8B, InternVL3-8B, and
InternVL3.5-8B~\citep{chen2025internvl25,chen2025internvl3,wang2025internvl35}.
For each video-capable backbone, we compare REZE with Direct timestamp
generation under the same evaluation protocol. The only exception is the
image-only LLaVA-1.5, which has no Direct setting and is therefore evaluated
only with REZE. To extend this same-backbone comparison beyond Direct, we also
include the training-free NumPro baseline~\citep{wu2025numpro} on Qwen2-VL-7B.
Beyond these controlled comparisons, we compare REZE against the reported
benchmark results of prior training-free and supervised grounding methods.

\begin{table*}[t]
\centering
{\renewcommand{\arraystretch}{1.12}\small
\begin{minipage}[t]{0.43\textwidth}
\vspace{0pt}
\centering
\textbf{(A) Same backbone, training-free}\par\smallskip
{\setlength{\tabcolsep}{3pt}
\begin{tabular}{@{}lrrr@{}}
\toprule
\textbf{Method} &
\textbf{\shortstack{Char.\\mIoU}} &
\textbf{\shortstack{HD\\mAP}} &
\textbf{HIT@1} \\
\midrule
Direct                 & 6.48  & 12.32 & 19.58 \\
NumPro (Qwen2-VL)     & 38.50 & 23.60 & 43.40 \\
\textbf{REZE}          & \textbf{51.49} & \textbf{43.28} & \textbf{71.66} \\
\bottomrule
\end{tabular}}

\medskip

\textbf{(C) Measured inference cost}\par\smallskip
{\setlength{\tabcolsep}{3pt}
\begin{tabular}{@{}lrrr@{}}
\toprule
\textbf{Method} &
\textbf{\shortstack{Input tokens\\per query}} &
\textbf{\shortstack{Throughput\\(queries/s)}} &
\textbf{\shortstack{Extra memory\\(GB/call)}} \\
\midrule
\multicolumn{4}{l}{\textit{Charades-STA}} \\
Direct & $\sim$5.6k & 0.41  & 0.5--1.7 \\
REZE   & 1.1k$\times$11 & 0.27  & 0.26 \\
\addlinespace[1pt]
\multicolumn{4}{l}{\textit{QVHighlights}} \\
Direct & $\sim$31.5k & 0.073 & 6.6--7.2 \\
REZE   & 1.4k$\times$51 & 0.052 & 0.31 \\
\bottomrule
\end{tabular}}
\end{minipage}\hfill
\begin{minipage}[t]{0.55\textwidth}
\vspace{0pt}
\centering
\textbf{(B) QVHighlights test results}\par\smallskip
{\setlength{\tabcolsep}{3pt}
\begin{tabular}{@{}lrrr@{}}
\toprule
\textbf{Method} & \textbf{MR} & \textbf{HD} & \textbf{HIT@1} \\
\midrule
\multicolumn{4}{l}{\textit{Target-benchmark training}} \\
CVA~\citep{cva}              & 47.49 & 44.43 & 66.01 \\
SG-DETR~\citep{sgdetr}       & 54.10 & 43.76 & 69.13 \\
HL-CLIP~\citep{hlclip}       & --    & 41.94 & 70.60 \\
CoSTL~\citep{costl}          & 46.68 & 41.36 & 71.59 \\
\multicolumn{4}{l}{\textit{External pretraining + target fine-tuning}} \\
SG-DETR (w/PT)~\citep{sgdetr} & \textbf{58.80} & \textbf{44.70} & 71.00 \\
\multicolumn{4}{l}{\textit{Training-free}} \\
GranAlign~\citep{granalign}  & 38.23 & --    & --    \\
NumPro (Qwen2-VL)~\citep{wu2025numpro} & -- & 23.60 & 43.40 \\
REZE (Qwen2.5-VL)            & \underline{40.32} & 42.21 & 69.58 \\
REZE (InternVL3)             & 39.33 & \underline{44.18} &
\textbf{\underline{73.41}} \\
\bottomrule
\end{tabular}}
\end{minipage}
}
\caption{Main performance and cost comparisons. (A)~Direct, NumPro, and REZE
use the same Qwen2-VL-7B backbone; columns report Charades-STA \miou{} and
QVHighlights highlight-detection mAP and HIT@1. (B)~Published QVHighlights
moment-retrieval (MR) mAP and highlight-detection (HD) mAP and HIT@1.
Underlining marks the best training-free result, while bold marks the best
overall result. Each REZE row contains one backbone. (C)~Measured cost on
Qwen2.5-VL-7B; input tokens are totals per query, throughput is measured in
queries per second, and extra memory is the transient GPU-memory increase per
call above the loaded model.
Percentages for all performance metrics.}
\label{tab:sota}
\end{table*}

\paragraph{Evaluation protocol.}
We use one fixed REZE configuration for each output type across all backbones.
Any data-dependent choices are made on the Charades-STA training split or the
QVHighlights validation split and then frozen before test evaluation. For
QVHighlights highlight detection, we convert the intervals predicted by Direct
into a binary mask over the two-second evaluation clips, while REZE uses its
clip-level scores directly. We evaluate the public QVHighlights test annotations
with the official Moment-DETR evaluator. Appendix~A
provides the complete evaluation and implementation details, including the
prompts, parsing rules, and test-release information.

\subsection{Main Results}
\label{sec:exp-main}

\paragraph{Controlled same-backbone comparison.}
With the evaluation protocol fixed, we first compare the three training-free
approaches on the same Qwen2-VL-7B backbone. As shown in
Table~\ref{tab:sota}A, REZE outperforms both Direct and NumPro when all three
use Qwen2-VL-7B. On Charades-STA, REZE reaches $51.49$ \miou{}, compared with
$6.48$ for Direct and $38.50$ for NumPro. For QVHighlights highlight detection, REZE
reaches $43.28$ mAP and $71.66$ HIT@1, whereas Direct reaches $12.32$ mAP and
$19.58$ HIT@1. With the same Qwen2-VL-7B backbone, NumPro improves these scores
to $23.60$ mAP and $43.40$ HIT@1 but remains below REZE. These results isolate
the improvement from the temporal grounding procedure rather than a stronger
backbone. Query-level bootstrap intervals and a paired
Direct--REZE test are reported in Appendix~B.

\paragraph{Training-free state of the art on QVHighlights.}
Having established the improvement on a fixed backbone, we next compare REZE
with published results on QVHighlights in Table~\ref{tab:sota}B. For moment
retrieval, REZE with Qwen2.5-VL-7B reaches $40.32$ mAP, improving the previous
best training-free result of $38.23$ reported by GranAlign~\citep{granalign}.
For highlight detection, REZE with InternVL3-8B establishes a new training-free
state of the art at $44.18$ mAP and $73.41$ HIT@1. SG-DETR retains a slightly
higher mAP of $44.70$, whereas REZE's $73.41$ HIT@1 surpasses CoSTL's $71.59$
and all other fully supervised SoTAs on the QVHighlights test
split~\citep{sgdetr,costl}.

\paragraph{Generality across task outputs.}
After evaluating multi-interval retrieval and highlight detection on
QVHighlights, we examine the same method on the single-interval benchmarks.
The controlled Charades-STA comparison in Table~\ref{tab:sota}A already shows
a substantial improvement. On ActivityNet Captions, REZE raises the \miou{} of
Qwen2.5-VL from $12.61$ to $36.47$, reaching the $36.76$ achieved by Qwen3-VL
through Direct timestamp generation. On Qwen3-VL itself, Direct remains slightly
higher than REZE ($36.76$ versus $35.43$ \miou{}). Qwen3-VL combines textual
timestamp alignment with large-scale spatio-temporal video grounding data, and
its post-training includes reinforcement learning on grounding
tasks~\citep{qwen3vl}. These components directly train the capability used by
Direct timestamp generation, which helps explain why moving temporal
aggregation outside the model no longer improves Qwen3-VL on ActivityNet.
Taken together, the results across all three output formats show that the same clip-level scoring formulation
supports single-interval retrieval, multi-interval retrieval, and highlight
detection by changing only the external readout. Appendix~B reports the complete per-benchmark results, including Recall.

\paragraph{Generality across backbones.}
Table~\ref{tab:crossbackbone} evaluates cross-backbone generalisation using
seven backbones on Charades-STA and four on QVHighlights. Across the
Charades-STA and QVHighlights results in this table, REZE improves over Direct
whenever a Direct baseline exists. On Charades-STA, Direct ranges from $6.48$
to $53.30$ \miou{} across six backbones, whereas REZE ranges from $42.37$ to
$55.66$ across all seven, including $42.37$ on the image-only LLaVA-1.5, which
has no Direct setting. Within the Qwen family, Qwen2-VL with REZE reaches
$51.49$ \miou{}, approaching the $53.30$ achieved by Qwen3-VL through Direct
timestamp generation.

\begin{table*}[t]
\centering
{\renewcommand{\arraystretch}{1.15}\small
\setlength{\tabcolsep}{3pt}
\begin{tabular}{l|ccc|ccc|ccc|ccc}
\toprule
 & \multicolumn{3}{c|}{\textbf{Charades \miou{}}} & \multicolumn{3}{c|}{\textbf{QVH MR mAP}} & \multicolumn{3}{c|}{\textbf{QVH HD mAP}} & \multicolumn{3}{c}{\textbf{QVH HD HIT@1}} \\
\cmidrule(lr){2-4}\cmidrule(lr){5-7}\cmidrule(lr){8-10}\cmidrule(lr){11-13}
\textbf{Backbone} & Direct & REZE & $\Delta$ & Direct & REZE & $\Delta$ & Direct & REZE & $\Delta$ & Direct & REZE & $\Delta$ \\
\midrule
LLaVA-1.5-7B   & --    & 42.37 & \textit{n/a} & --    & --    & --    & --    & --    & --    & --    & --    & --    \\
InternVL2.5-8B & 9.35  & 50.93 & $+41.6$      & --    & --    & --    & --    & --    & --    & --    & --    & --    \\
InternVL3-8B   & 20.16 & 52.91 & $+32.8$      & 9.67  & 39.33 & $+29.7$ & 16.49 & \textbf{44.18} & $+27.7$ & 25.29 & \textbf{73.41} & $+48.1$ \\
InternVL3.5-8B & 30.17 & 52.70 & $+22.5$      & --    & --    & --    & --    & --    & --    & --    & --    & --    \\
Qwen2-VL-7B    & 6.48  & 51.49 & $+45.0$      & 6.20  & 34.99 & $+28.8$ & 12.32 & 43.28 & $+31.0$ & 19.58 & 71.66 & $+52.1$ \\
Qwen2.5-VL-7B  & 42.74 & 54.22 & $+11.5$      & 32.11 & \textbf{40.32} & $+8.2$ & 24.03 & 42.21 & $+18.2$ & 33.07 & 69.58 & $+36.5$ \\
Qwen3-VL-8B    & 53.30 & 55.66 & $+2.4$       & 26.60 & 28.21 & $+1.6$  & 22.98 & 43.37 & $+20.4$ & 35.21 & 71.98 & $+36.8$ \\
\bottomrule
\end{tabular}}
\caption{Per-backbone Direct versus REZE under one protocol. Charades-STA
\miou{}; QVHighlights \texttt{test} moment retrieval (MR mAP) and highlight
detection (HD mAP, HIT@1). $\Delta$ is REZE minus
Direct. REZE improves on Direct wherever Direct is defined, and the gain shrinks
as the backbone's own generation matures. Image-only LLaVA-1.5 has no native
video-time interface, so no comparable Direct gap is reported (\textit{n/a}): REZE
enables this backbone rather than improving a baseline. Direct HD is a naive
binary mask (compare against trained detectors, Table~\ref{tab:sota}). ``--'':
undefined or not measured. Full Recall in Appendix~B.}
\label{tab:crossbackbone}
\end{table*}

\paragraph{Accuracy--cost trade-off.}
Table~\ref{tab:sota}C places the corresponding computational cost beside the
main performance comparisons. On Qwen2.5-VL-7B, REZE processes
$2.2$--$2.3\times$ more input tokens per query, while Direct achieves
$1.4$--$1.5\times$ higher batched query throughput. In exchange, each REZE call contains only
$1.1$--$1.4$k input tokens and uses less than $0.35$\,GB of transient memory,
while a Direct call on a $150$-second QVHighlights video reaches approximately
$31.5$k tokens and $7$\,GB. REZE therefore trades total throughput for a bounded
per-call context and memory footprint. Appendix~E
provides the complete measurement protocol and per-dataset results.

\subsection{Ablations and Analysis}
\label{sec:exp-analysis}

\begin{figure}[t]
\centering
\includegraphics[width=\columnwidth]{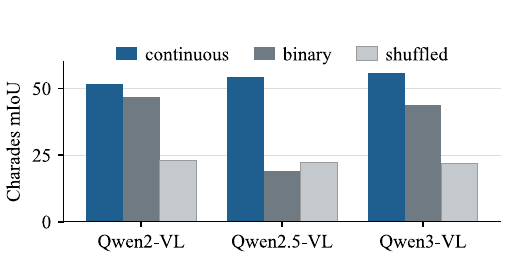}
\caption{Score representation on Charades-STA. Continuous clip scores are
compared with hard binary decisions and the same continuous scores shuffled
along the video timeline across three Qwen backbones. Percentages.}
\label{fig:analysis}
\end{figure}

\paragraph{Continuous and temporally ordered scores.}
We first examine whether the ordered continuous score sequence provides useful
temporal evidence. Figure~\ref{fig:analysis} compares the default REZE scores
with two controls. Binary keeps the same clips, prompt, and external aggregation
as REZE, but replaces the continuous score obtained by normalising the
\texttt{Yes} and \texttt{No} logits with a hard decision: the clip receives $1$
when \texttt{Yes} has the larger logit and $0$ otherwise. Shuffled keeps the
continuous scores unchanged but randomly permutes their positions along the
video timeline. Continuous scores achieve the highest \miou{} on all three Qwen
backbones. The effect is largest for Qwen2.5-VL, where binarisation reduces
\miou{} from $54.22$ to $19.10$, while shuffling reduces all three backbones to
$21.8$--$23.1$. Because shuffling preserves the score values but removes their
temporal positions, this decline shows that REZE uses temporally ordered
evidence and cannot be explained by the score distribution or a duration
preference alone~\citep{otani2020}.

\paragraph{Single-interval extractor.}
REZE uses the mean-shifted max-subarray defined in the Method section. To test
whether its performance depends on this particular extractor,
Table~\ref{tab:ablation}A replaces only the extractor while keeping the saved
Qwen3-VL score curves unchanged. This sensitivity analysis is performed on the
official Charades-STA training split, while the extractor used for the reported
test results remains fixed. The mean-shifted max-subarray reaches $53.49$
\miou{}, while the cumulative-sum scan and its standardised variant reach
$53.41$ and $53.21$. Their similar results show that performance does not
depend on selecting one particular scanning algorithm. By contrast,
threshold-based extractors trail by $6$ to $14$ points, indicating that the
important design choice is to accumulate evidence relative to the video mean.
We use the same max-subarray for all test results because it returns exactly one
interval without introducing an additional threshold.

\begin{table}[t]
\centering
{\renewcommand{\arraystretch}{1.15}\small
\setlength{\tabcolsep}{3pt}
\begin{tabular}{lc}
\toprule
\multicolumn{2}{l}{\textit{(A) Single-interval extractor (Charades-STA \texttt{train}, \miou{})}} \\
\midrule
mean-shifted max-subarray \textit{(used)} & \textbf{53.49} \\
cumulative-sum scan & 53.41 \\
standardised cumulative-sum & 53.21 \\
median-shifted max-subarray & 45.24 \\
percentile (60/70/80th) & 47.35/46.48/42.89 \\
adaptive mean${}+k\sigma$ ($k{=}0.3/0.5/1$) & 45.44/44.45/39.02 \\
\midrule
\multicolumn{2}{l}{\textit{(B) Multi-interval aggregation (QVH val, MR mAP)}} \\
\midrule
mean threshold, $\sigma{=}\lfloor D\rfloor/15$ & 20.76 \\
mean threshold, $\sigma{=}3$\,s & 36.77 \\
mean threshold, $\sigma{=}2$\,s & 39.74 \\
Otsu threshold, $\sigma{=}3$\,s \textit{(used)} & \textbf{42.46} \\
\bottomrule
\end{tabular}}
\caption{Sensitivity of the external readout. Panel A changes only the
single-interval extractor using the same saved Qwen3-VL score curves. Panel B
changes the smoothing and threshold for QVHighlights multi-interval retrieval
using Qwen2.5-VL. Percentages.}
\label{tab:ablation}
\end{table}

\paragraph{Multi-interval aggregation.}
The multi-interval readout has a different requirement because it must separate
several relevant regions in a long video. Table~\ref{tab:ablation}B evaluates
the smoothing bandwidth and threshold on the QVHighlights validation split. To
separate their effects, the first three settings use the same per-video mean
threshold and vary only the smoothing bandwidth. The duration-dependent
bandwidth $\sigma=\lfloor D\rfloor/15$, where $D$ is the video duration, reaches
$20.76$ mAP. Replacing it with a fixed three-second bandwidth, matched to the
clip duration, raises mAP to $36.77$, while a fixed two-second bandwidth reaches
$39.74$. The two fixed settings show that the improvement comes from preventing
the smoothing bandwidth from growing with the video duration, rather than from
the exact choice of three seconds. Finally, with the three-second bandwidth held
fixed, replacing the video-mean threshold with the per-curve Otsu threshold
further raises mAP to $42.46$. This contrast explains why the multi-interval
readout uses a short fixed bandwidth, while the single-interval readout uses a
duration-dependent bandwidth.

\paragraph{Verifier strength and boundary precision.}
The aggregation study resolves the dependence on video length, but it does not
remove the resolution limit of one score per clip. In the short-moment
diagnostic reported in Appendix~C, Qwen3-VL has the
largest median ratio between scores on relevant and background clips, yet it
reaches only $28.21$ moment-retrieval mAP on QVHighlights, below Qwen2-VL's
$34.99$. A sharper verifier therefore does not by itself recover sub-clip
boundary precision: clip-level recognition supports highlight ranking, while
short-moment retrieval additionally requires evidence of where an event begins
and ends within a clip.

\section{Limitations}
\label{sec:limitations}

REZE scores each short clip independently and therefore does not explicitly
model relations across clips. Tasks that require temporal ordering, event
counting, or motion direction fall outside the current design. In addition, the
ActivityNet evaluation covers only the Qwen family, and the Qwen3-VL result in
Section~\ref{sec:exp-main} shows that REZE does not improve every
backbone--dataset pair. Appendix~G provides further
details.

REZE also trades query-level efficiency for a smaller per-call memory
footprint. On Qwen2.5-VL-7B, it processes $2.2$--$2.3\times$ as many input
tokens per query as Direct, while Direct achieves $1.4$--$1.5\times$ higher
batched query throughput. In return, Direct uses $1.9$--$6.5\times$ as much
transient memory per call on Charades-STA and $21$--$23\times$ as much on
QVHighlights (Table~\ref{tab:sota}C).

\section{Conclusion}
\label{sec:conclusion}

REZE performs video temporal grounding without task-specific training and
without asking the VLM to generate timestamps. It obtains a confidence score
for the query from each short clip and constructs the required temporal output
through external aggregation. By changing only the aggregation rule, the same
clip-level scoring procedure supports single-interval retrieval, multi-interval
retrieval, and highlight detection. Since the procedure only requires
clip-level scores, it applies to both video-capable and image-only backbones.
Across seven backbones from three model families, REZE improves every available
Direct comparison on Charades-STA and QVHighlights and establishes new
training-free state-of-the-art results on both QVHighlights tasks. Together,
these results show that the clip-level recognition ability already available
in VLMs can serve as a common basis for different VTG tasks, while external
aggregation adapts the scores to each required output.

\bibliography{references}

\clearpage

\appendix
\setcounter{secnumdepth}{0}

\maketitle

\section{Appendices}

\subsection{A. Implementation and Reproducibility}
\label{app:code}

This appendix gives the core \reze{} code, the REZE and Direct prompts, the output format,
and the design details that the main text only summarises. All code
is original and relies on the following third-party libraries, used as released:
PyTorch, vLLM, HuggingFace Transformers, \texttt{qwen-vl-utils}, OpenCV, NumPy, and
SciPy. Charades-STA, ActivityNet Captions, and QVHighlights are public datasets
used as released, and all backbones are publicly released pretrained weights.
As in the main paper, \emph{Direct} denotes direct timestamp generation.
All headline accuracy results use one deterministic greedy-decoding pass per
query. Randomised ancillary analyses use fixed seeds: $0$ for bootstrap and
efficiency sampling, and $2027$ for validation-subset construction.
The experiments were run on Ubuntu 22.04 with a single NVIDIA RTX 6000 Ada
Generation GPU with 48\,GB of memory. The archived environment records Python
3.10.19, PyTorch 2.10.0 with CUDA 12.9, NumPy 2.2.6, and SciPy 1.15.3.

Videos are divided from the start into consecutive clips of at most $3$ seconds.
A final residual shorter than $0.3$ seconds is discarded, and every retained
clip is sampled at $2$ frames per second. The interval extractor below is the
single-interval readout. For QVHighlights moment retrieval, REZE instead smooths
the per-second curve with $\sigma=3$ seconds and applies an Otsu threshold.
Contiguous runs above the threshold are candidate intervals; runs shorter than
$2$ seconds are removed, and the remaining candidates are ranked by their
accumulated excess above the threshold before the top ten are returned. If no
candidate survives, the whole video is returned as one interval. Highlight
detection uses no smoothing or thresholding: it averages the raw per-second
curve over the benchmark's consecutive $2$-second clips.

\subsubsection{A.1. Direct timestamp-generation prompts}

Direct uses no system prompt. For the single-interval Charades-STA and
ActivityNet Captions benchmarks, the user prompt is:

\begin{lstlisting}
Please find the visual event described by the sentence '{query}', determining its starting and ending times. The format should be: 'The event happens in <start time> - <end time> seconds'.
\end{lstlisting}

Because QVHighlights moment retrieval can contain several relevant intervals,
Direct instead receives the following duration-aware prompt:

\begin{lstlisting}
You are given a clipped video. Its duration is {duration:.2f} seconds.
All timestamps must be relative to THIS clipped video.
The valid timestamp range is from 0.00 to {duration:.2f} seconds.

Task: find all time windows where the following description occurs:
"{query}"

Rules:
1. Only output timestamps inside [0.00, {duration:.2f}].
2. If the event appears multiple times, list every separate window.
3. If you are unsure, output the most likely window.
4. Do not include explanations.

Output format, one window per line:
[start_seconds, end_seconds]
\end{lstlisting}

The Qwen2-VL QVHighlights implementation further clarifies that timestamps
refer to the clipped video rather than the original source video; the remaining
instructions are unchanged. All Direct responses use greedy decoding
(\texttt{temperature=0}), with at most $128$ output tokens, or $256$ for this
Qwen2-VL implementation. Thinking mode is disabled for Qwen3-VL.

\subsubsection{A.2. Evaluation release and Direct parsing}

We evaluate QVHighlights on the public held-out test labels released in the
Moment-DETR repository after the CodaLab server was closed (commit
\texttt{7c61a84}, 9 March 2026). Predictions are matched to
\texttt{highlight\_test\_release} by query identifier and scored with the
released Moment-DETR evaluator.

For the single-interval benchmarks, the Direct parser extracts the first two
non-negative numbers from the response, orders them as start and end times, and
returns a parse failure when fewer than two numbers are present or the two
times are equal. Every parsed interval is scored by its IoU; only parse failures
and zero-length intervals receive zero by definition. For QVHighlights moment
retrieval, the parser accepts bracketed pairs, ``from \ldots{} to \ldots{}''
expressions, and two numbers separated by ``to'', a hyphen, or a comma. It
orders each pair, clips both endpoints to the video duration, removes intervals
shorter than $0.3$ seconds, removes near-duplicate pairs within $0.5$ seconds,
and keeps at most ten predictions. For highlight detection, each predicted
Direct interval becomes a binary mask over the benchmark's two-second
evaluation clips.

\subsubsection{A.3. Per-clip verification prompt}

\begin{lstlisting}[language=Python]
SYSTEM = ("You are a strict video action verifier. Your default answer is "
          "'No'. Answer 'Yes' ONLY if you are highly confident the described "
          "action is clearly and actively occurring in the frames. If there "
          "is any doubt, answer 'No'. Output: a single word 'Yes' or 'No'.")

def build_user(query, n_frames):
    return f"Action: {query}\nIs this action CLEARLY occurring in these " \
           f"{n_frames} frames? Answer:"
\end{lstlisting}

The prompt is deliberately asymmetric: the model defaults to \texttt{No} and commits
to \texttt{Yes} only when the action is clearly present, which biases the verifier
against background false positives and widens the $P(\text{Yes})$ gap.

\subsubsection{A.4. Per-clip aggregation and interval extraction}

Every verifier call is issued with greedy decoding, a single output token, and the
top-20 next-token log-probabilities (\texttt{max\_tokens=1}, \texttt{logprobs=20},
\texttt{temperature=0}); the score of Equation~1 is then computed from the
returned log-probabilities as below. The final branch of \texttt{extract\_pyes} is
a fallback for the rare case where one answer token falls outside the returned
top-20, which happens only at extreme confidence. Rather than failing the query,
the score is then approximated by the remaining token's probability, or set to
$0$ when \texttt{Yes} itself is absent. Because the prompt restricts the answer to
\texttt{Yes} or \texttt{No}, both tokens are almost always inside the top-20, so
this branch rarely fires and leaves the reported curves unchanged. The ablations
of Appendix~C show that the pipeline does not rely on fine
calibration of the score in any case.

\begin{lstlisting}[language=Python]
import math, re, numpy as np
from scipy.ndimage import gaussian_filter1d
from scipy.interpolate import interp1d

# vLLM call: SamplingParams(max_tokens=1, logprobs=20, temperature=0.0)
def extract_pyes(logprobs, yes_id, no_id):
    y, n = logprobs.get(yes_id), logprobs.get(no_id)
    if y is not None and n is not None:       # normal path: Eq. (1)
        m = max(y.logprob, n.logprob)
        ey, en = math.exp(y.logprob - m), math.exp(n.logprob - m)
        return ey / (ey + en)
    # fallback: one token left the top-20 (extreme confidence only)
    return math.exp(y.logprob) if y is not None else 0.0

def segments_to_curve(scores, segments, duration, resolution=1.0):
    n = max(1, int(duration / resolution))
    curve, counts = np.zeros(n), np.zeros(n)
    for sc, (s, e) in zip(scores, segments):
        for i in range(max(0, int(s / resolution)), min(n, int(e / resolution) + 1)):
            curve[i] += sc; counts[i] += 1
    mask = counts > 0
    curve[mask] /= counts[mask]
    if not mask.all() and mask.any():
        vi = np.where(mask)[0]
        curve = interp1d(vi, curve[vi], kind='nearest',
                         fill_value='extrapolate')(np.arange(n)) if len(vi) > 1 else \
                np.full(n, curve[vi[0]])
    return curve

def maxsubarray_extract(curve, duration):
    n = len(curve)  # per-second curve length, i.e. floor(D)
    sm = gaussian_filter1d(curve, sigma=max(1.0, n / 15.0))  # sigma = floor(D)/15
    shifted = sm - sm.mean()
    best, bs, be, cs, cstart = -float('inf'), 0, 0, 0.0, 0
    for i in range(n):
        cs += shifted[i]
        if cs > best:
            best, bs, be = cs, cstart, i
        if cs < 0:
            cs, cstart = 0.0, i + 1
    return (bs / n) * duration, ((be + 1) / n) * duration
\end{lstlisting}

Two discretisation conventions are visible in the listing: a boundary second
shared by two adjacent clips receives their average, and the per-second curve has
$\lfloor D\rfloor$ integer-second positions that are mapped back to the full duration. Re-running afterwards
with half-open bin assignment and unrescaled one-second bins shifts
Charades \miou{} by between $-1.5$ and $+0.1$ across the Qwen backbones, so
neither convention drives the results.

\subsubsection{A.5. Design choices (full)}

\paragraph{Why segment-level rather than frame-level?} We score short clips
rather than individual frames for two reasons. First, a single frame lacks the
motion context that differentiates ``picking up the cup'' from ``putting down
the cup''. Second, calling the VLM once per sampled frame would require roughly
six times as many forward passes under our six-frame clip construction. A
three-second clip therefore balances short-motion context against call count
and produces one decision for each local portion of the video. This concerns
the scale at which we decide, not how each clip is scored internally. A
multi-frame backbone can exploit the within-clip motion jointly, while an
image-only backbone uses the same six-frame temporal support but aggregates
independent frame-wise evidence into one clip-level decision. Both therefore
decide at clip rather than frame level.

\paragraph{Why externalise temporal aggregation?} Direct generation requires the
VLM to (a) recognise the action wherever it occurs, (b) maintain a consistent
absolute-time coordinate, and (c) verbalise a parseable string. Delegating (b) and
(c) to a deterministic algorithm isolates (a), the local recognition signal
measured by the clip-level verifier. The diagnostics of
Appendix~D show that failures at these two stages account for a
large share of the error when Direct performs poorly.

\paragraph{Additional extractor check.} Table~3 in the main paper compares the
mean-shifted max-subarray with nine alternative single-interval extractors. We
also checked afterwards whether the multi-interval Otsu readout could serve
here by retaining only its highest-scoring interval. On Charades-STA, it remains
within one \miou{} point of the max-subarray on all three Qwen backbones. The
comparison script is released with the code.

\subsubsection{A.6. Output format}
Each run writes one JSON record per query (here \reze{} on Qwen2-VL-7B, Charades-STA;
\texttt{scores} abbreviated):

\begin{lstlisting}[language=Python]
{"id": "55NRK_1", "video": "55NRK.mp4",
 "query": "person begins sneezing.",
 "scores": [0.22, 0.55, 0.53, 0.55, 0.42, 0.59, 0.71, 0.84, ...],
 "segments": [[0.0, 3.0], [3.0, 6.0], ..., [30.0, 30.6]],
 "pred_start": 19.4, "pred_end": 30.6, "duration": 30.6,
 "gt_start": 19.1, "gt_end": 30.6, "iou": 0.98}
\end{lstlisting}

\subsection{B. Complete Main Results}
\label{app:results}

\paragraph{Statistical uncertainty.}
A query-level bootstrap with $2{,}000$ resamples gives $95\%$ confidence
intervals of $[54.7,56.6]$ \miou{} on Charades-STA (Qwen3-VL),
$[38.5,42.2]$ mAP on QVHighlights moment retrieval (Qwen2.5-VL), and
$[43.0,45.3]$ mAP and $[71.3,75.6]$ HIT@1 on highlight detection
(InternVL3). Because the per-query predictions of prior methods are not
available, these intervals characterise uncertainty in the REZE estimates
rather than a paired comparison with published methods. A paired Wilcoxon
signed-rank test between REZE and Direct on Qwen2-VL Charades-STA gives
$p<0.001$.

\subsubsection{B.1. Full Cross-Benchmark Leaderboard}
The compact comparison of Table~1 in the main paper is expanded in
Table~\ref{tab:sota-full} with the full competitor list, grouped by training
regime. Each REZE row contains one backbone, while composite baselines are
explicitly labelled; the four backbones evaluated on QVHighlights each appear
with their own Charades and QVHighlights numbers. The $\dagger$ marks methods built on
InternVideo2 features, and highlight detection varies only modestly across the
REZE backbones ($42$--$44$ mAP and $70$--$73$ HIT@1). We omit target-trained
Charades-STA detectors and do not claim a Charades state of the art: REZE's
highest Charades number sits on the strongest backbone (Qwen3-VL), where it
roughly matches that backbone's own native generation ($53.30$); REZE's value on
Charades lies instead in the large gain it brings to weaker backbones and to
image-only LLaVA-1.5 (Table~\ref{tab:charades-main}).

\begin{table*}[t]
\centering
{\renewcommand{\arraystretch}{1.12}\small
\setlength{\tabcolsep}{6pt}
\begin{tabular}{lcccc}
\toprule
 & \textbf{Charades} & \textbf{QVH-MR} & \textbf{QVH-HD} & \textbf{QVH-HD} \\
\textbf{Method} & mIoU & mAP & mAP & HIT@1 \\
\midrule
\multicolumn{5}{l}{\textit{Trained on the target benchmark}} \\
Moment-DETR \citep{lei2021qvhighlights} & --    & 30.73 & 35.69 & 55.60 \\
HL-CLIP \citep{hlclip}               & --    & --    & 41.94 & 70.60 \\
CVA \citep{cva}                      & --    & 47.49 & 44.43 & 66.01 \\
SDST$^\dagger$ \citep{sdst}          & --    & 53.31 & 43.40 & 69.13 \\
SG-DETR$^\dagger$ \citep{sgdetr}     & --    & 54.10 & 43.76 & 69.13 \\
CoSTL \citep{costl}                  & --    & 46.68 & 41.36 & 71.59 \\
\midrule
\multicolumn{5}{l}{\textit{External pretraining + target fine-tuning}} \\
SG-DETR (w/PT)$^\dagger$ \citep{sgdetr} & -- & \textbf{58.80} & \textbf{44.70} & 71.00 \\
\midrule
\multicolumn{5}{l}{\textit{Externally pretrained, zero-shot on target}} \\
SG-DETR (ZS) \citep{sgdetr}          & --    & 48.30 & 43.00 & 68.00 \\
\midrule
\multicolumn{5}{l}{\textit{Training-free (no gradient updates)}} \\
Robust Zero-Shot (3-VLM ens.) \citep{robustzs} & 47.95 & --    & --    & --    \\
NumPro (Qwen2-VL-7B) \citep{wu2025numpro}  & 38.50 & --    & 23.60 & 43.40 \\
GranAlign \citep{granalign}                & 38.00 & 38.23 & --    & --    \\
REZE, Qwen2-VL-7B \textit{(ours)}    & 51.49 & 34.99 & 43.28 & 71.66 \\
REZE, Qwen2.5-VL-7B \textit{(ours)}  & 54.22 & 40.32 & 42.21 & 69.58 \\
REZE, Qwen3-VL-8B \textit{(ours)}    & 55.66 & 28.21 & 43.37 & 71.98 \\
REZE, InternVL3-8B \textit{(ours)}   & 52.91 & 39.33 & 44.18 & \textbf{73.41} \\
\bottomrule
\end{tabular}}
\caption{Complete cross-benchmark comparison, grouped by how much training each method uses
(expanded form of Table~1 in the main paper). Charades-STA \miou{}; QVHighlights
\texttt{test} moment-retrieval mAP and highlight-detection mAP and HIT@1. REZE is
listed once per backbone, so every REZE row contains one model; composite
baselines are explicitly labelled.
$\dagger$ marks methods built on InternVideo2 features; ``--'' marks a benchmark a
method does not report. Bold marks the best value in each QVHighlights column
(the Charades column is not a target of this work; see text). Percentages.}
\label{tab:sota-full}
\end{table*}

\subsubsection{B.2. Charades-STA}
Table~\ref{tab:charades-main} reports all seven backbones. REZE's minimum is
$42.37$ (LLaVA-1.5-7B) and maximum $55.66$ (Qwen3-VL-8B), a $13.3$-point band,
while Direct rises from non-existent (image-only) through $6.48$ (Qwen2-VL), $42.74$
(Qwen2.5-VL), to $53.30$ (Qwen3-VL). The same pattern, REZE nearly flat while Direct catches
up, holds within each family: in Qwen, REZE goes $51.5 \to 55.7$ while Direct
goes $6.5 \to 53.3$; in InternVL, REZE goes $50.9 \to 52.7$ while Direct goes
$9.4 \to 30.2$.

\begin{table*}[t]
\centering
{\renewcommand{\arraystretch}{1.1}\small
\setlength{\tabcolsep}{5pt}
\begin{tabular}{lcccc}
\toprule
\textbf{Model} & \textbf{\miou} & \textbf{R@0.3} & \textbf{R@0.5} & \textbf{R@0.7} \\
\midrule
\multicolumn{5}{l}{\textit{Image-only backbone}} \\
LLaVA-1.5-7B     & --    & --    & --    & --    \\
\quad\textit{+\,REZE} & 42.37 & 63.20 & 41.02 & 20.86 \\
\midrule
\multicolumn{5}{l}{\textit{InternVL family}} \\
InternVL2.5-8B & 9.35  & 10.56 & 5.16  & 2.61  \\
\quad\textit{+\,REZE} & 50.93 & 75.46 & 53.63 & 29.27 \\
InternVL3-8B   & 20.16 & 27.74 & 17.66 & 8.90  \\
\quad\textit{+\,REZE} & 52.91 & 77.23 & 58.82 & 33.33 \\
InternVL3.5-8B & 30.17 & 44.84 & 28.25 & 12.72 \\
\quad\textit{+\,REZE} & 52.70 & 78.39 & 58.09 & 31.75 \\
\midrule
\multicolumn{5}{l}{\textit{Qwen-VL family}} \\
Qwen2-VL-7B    & 6.48  & 6.88  & 4.81  & 2.15  \\
\quad\textit{+\,REZE} & 51.49 & 77.63 & 54.62 & 29.46 \\
Qwen2.5-VL-7B  & 42.74 & 63.66 & 43.17 & 22.66 \\
\quad\textit{+\,REZE} & 54.22 & 79.78 & 59.92 & 35.05 \\
Qwen3-VL-8B    & 53.30 & 79.87 & 62.88 & 32.15 \\
\quad\textit{+\,REZE} & \textbf{55.66} & \textbf{80.59} & \textbf{64.92} & \textbf{37.80} \\
\bottomrule
\end{tabular}}
\caption{Main results on Charades-STA test (N=3720). Each backbone appears as
its Direct timestamp generation (model row) and the same model under
REZE (the indented \textit{+\,REZE} rows). REZE \miou{} stays within a band of
about $13.3$ points (lowest $42.37$ on LLaVA-1.5-7B, highest $55.66$ on
Qwen3-VL-8B), whereas Direct ranges from undefined on the image-only backbone
up to $53.30$. ``--''
marks models whose direct timestamp generation is unreliable or undefined. Bold
marks the best value in each column. All values in percentages.}
\label{tab:charades-main}
\end{table*}

\subsubsection{B.3. ActivityNet}
On the longer ActivityNet videos (Table~\ref{tab:anet-main}) REZE reaches about
$35$ \miou{}, over $15$ points below Charades, as a local six-frame verifier is
less useful when the action can span a quarter of a three-minute video. Within
Qwen, Direct is essentially flat from Qwen2-VL ($12.83$) to Qwen2.5-VL ($12.61$)
and jumps only on Qwen3-VL ($36.76$). Qwen3-VL is the one case where Direct ($36.76$) slightly
exceeds REZE ($35.43$). Table~\ref{tab:sigma-single} reports the corresponding
single-interval bandwidth sweep across all three Qwen backbones.

\begin{table}[t]
\centering
{\renewcommand{\arraystretch}{1.1}\small
\setlength{\tabcolsep}{6pt}
\begin{tabular}{lcccc}
\toprule
\textbf{Model} & \textbf{\miou} & \textbf{R@0.3} & \textbf{R@0.5} & \textbf{R@0.7} \\
\midrule
Qwen2-VL-7B    & 12.83 & 17.23 & 9.75  & 4.87  \\
\quad\textit{+\,REZE} & 35.74 & 53.11 & 33.10 & 15.23 \\
Qwen2.5-VL-7B  & 12.61 & 16.41 & 8.75  & 4.20  \\
\quad\textit{+\,REZE} & 36.47 & \textbf{54.64} & 33.85 & 15.51 \\
Qwen3-VL-8B    & \textbf{36.76} & 52.93 & \textbf{33.94} & \textbf{18.67} \\
\quad\textit{+\,REZE} & 35.43 & 53.10 & 32.80 & 15.04 \\
\bottomrule
\end{tabular}}
\caption{Main results on ActivityNet val\_2 (N=17031), shown as Direct
(model row) and \textit{+\,REZE} (indented), as in Table~\ref{tab:charades-main}.
InternVL ActivityNet evaluation is deferred to future work. Bold marks the best
value in each column. All values in percentages.}
\label{tab:anet-main}
\end{table}

\subsubsection{B.4. QVHighlights, per backbone}
\begin{figure}[t]
\centering
\includegraphics[width=\columnwidth]{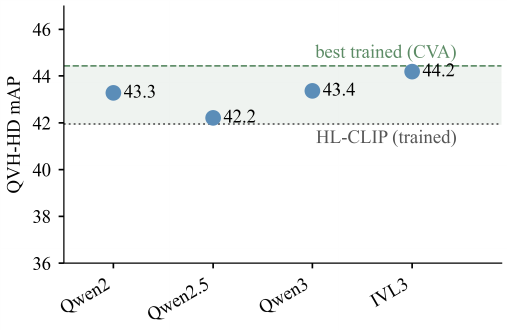}
\caption{QVHighlights \texttt{test} highlight-detection mAP of REZE on four
backbones (dots) against selected target-trained baselines (shaded, HL-CLIP to
CVA). All four sit inside this band. Test $N{=}1542$.}
\label{fig:hd-range}
\end{figure}

Table~2 of the main paper gives the four-backbone Direct-versus-REZE
breakdown, and Figure~\ref{fig:hd-range} plots the highlight-detection band
against the fully trained range. The margin over Direct on moment retrieval is
smallest on Qwen3-VL ($+1.6$ mAP): its verifier curves are highly polarised
(near $0$ or $1$) and fragment long intervals under a fixed small bandwidth. A
per-model aggregator could raise this case further, but the reported numbers
keep the single Otsu aggregation for all four backbones to avoid per-model tuning.
InternVL3-8B direct generation uses a sparse $40$-frame input, the most its
$12{,}288$-token context admits for a $150$-second video, whereas REZE issues
one short clip per call.

\subsection{C. Additional Ablations}
\label{app:ablation}

\paragraph{Continuous scores and temporally shuffled scores.}
Hard-thresholding $p_i$ to $\{0,1\}$ drops Qwen2.5-VL \miou{} from $54.22$ to
$19.10$ on Charades and $36.47$ to $15.48$ on ActivityNet, showing that the
continuous scores carry information the hard decision loses. Shuffling segment scores in time drops
\miou{} to $21$--$25$ for all six dataset/model pairs (the temporally shuffled
null; Charades values are the mean over ten permutations, standard deviation
below $0.5$). Because shuffling preserves the score distribution but destroys
its temporal order, REZE's $11$--$35$-point margin shows that its gains cannot
be explained by a learned duration heuristic alone~\citep{otani2020}.
Verifier discriminability (GT/non-GT ratio) rises
$1.91\times \to 4.28\times$ from Qwen2-VL to Qwen3-VL on Charades while REZE \miou{}
barely moves (Table~\ref{tab:score-dist}), matching the diminishing returns of
verifier strength discussed in Section~4.3 of the main paper.

\begin{table*}[t]
\centering
{\small
\begin{tabular}{llccccccc}
\toprule
\textbf{Dataset} & \textbf{Model} & \textbf{Logits} & \textbf{Binary} & \textbf{Shuffled} & \textbf{GT mean} & \textbf{Non-GT mean} & \textbf{Ratio} & \textbf{FP\textsubscript{>0.3}} \\
\midrule
\multirow{3}{*}{Charades-STA}
  & Qwen2-VL-7B    & 51.49 & 46.93 & 23.08 & 0.700 & 0.366 & 1.91$\times$ & 45.6\% \\
  & Qwen2.5-VL-7B  & 54.22 & 19.10 & 22.21 & 0.247 & 0.099 & 2.50$\times$ & 7.5\%  \\
  & Qwen3-VL-8B    & 55.66 & 43.62 & 21.82 & 0.453 & 0.106 & 4.28$\times$ & 12.0\% \\
\midrule
\multirow{3}{*}{ActivityNet}
  & Qwen2-VL-7B    & 35.74 & 30.26 & 24.65 & 0.710 & 0.524 & 1.35$\times$ & 64.2\% \\
  & Qwen2.5-VL-7B  & 36.47 & 15.48 & 23.48 & 0.228 & 0.126 & 1.81$\times$ & 14.9\% \\
  & Qwen3-VL-8B    & 35.43 & 26.00 & 22.52 & 0.360 & 0.180 & 2.00$\times$ & 19.7\% \\
\bottomrule
\end{tabular}}
\caption{Score distribution analysis across three Qwen models and two
datasets. ``Logits'' is the default REZE using continuous P(Yes);
``Binary'' replaces continuous scores with hard 0/1 thresholding;
``Shuffled'' applies REZE to randomly permuted segment scores
(duration-prior baseline). GT/Non-GT mean denotes the average P(Yes)
on segments inside / outside ground-truth intervals; Ratio is GT/Non-GT.
FP\textsubscript{>0.3} is the fraction of non-GT segments scoring above 0.3.
On Charades-STA, the Shuffled column is the mean over ten random permutations (standard deviation below $0.5$). All \miou{} values in percentages.}
\label{tab:score-dist}
\end{table*}

\paragraph{Single-interval smoothing bandwidth.}
Table~\ref{tab:sigma-single} varies only the Gaussian bandwidth while holding
the clips, prompt, saved confidence scores, mean shift, and max-subarray
extractor fixed. No fixed bandwidth is best across all three ActivityNet
backbones. We therefore retain one duration-dependent rule for both
single-interval datasets instead of selecting a bandwidth for each dataset or
backbone.

\begin{table}[t]
\centering
{\renewcommand{\arraystretch}{1.1}\small
\setlength{\tabcolsep}{4pt}
\begin{tabular}{lcccc}
\toprule
\multicolumn{5}{l}{\textit{(A) Charades-STA}} \\
\midrule
\textbf{Bandwidth} & \textbf{Train} & \textbf{Test} & \textbf{Min} & \textbf{Max} \\
\midrule
$\max(1,\lfloor D\rfloor/15)$ \textit{(used)} & 53.41 & 51.47 & 42.37 & 55.66 \\
$2$\,s & 53.82 & 51.70 & 42.48 & 56.15 \\
$3$\,s & 52.70 & 50.64 & 42.21 & 54.70 \\
$5$\,s & 49.06 & 47.60 & 41.03 & 50.57 \\
no smoothing & 47.96 & 50.06 & 41.84 & 52.99 \\
\bottomrule
\end{tabular}
\par\medskip
\begin{tabular}{lccc}
\toprule
\multicolumn{4}{l}{\textit{(B) ActivityNet Captions}} \\
\midrule
\textbf{Bandwidth} & \textbf{Qwen2-VL} & \textbf{Qwen2.5-VL} & \textbf{Qwen3-VL} \\
\midrule
$\max(1,\lfloor D\rfloor/15)$ \textit{(used)} & 35.73 & 36.46 & 35.26 \\
$2$\,s & 36.36 & 36.67 & 33.75 \\
$3$\,s & 36.24 & 36.78 & 34.25 \\
$5$\,s & 36.05 & 36.73 & 34.79 \\
no smoothing & 36.37 & 36.33 & 31.86 \\
\bottomrule
\end{tabular}}
\caption{Single-interval Gaussian-bandwidth sensitivity (\miou{}).
(A)~Charades-STA uses Qwen3-VL on the official training split
($N{=}12{,}408$); Test, Min, and Max summarise the seven reported backbones on
the test split ($N{=}3{,}720$ each). (B)~ActivityNet Captions contains
$17{,}031$ queries per backbone. Only $\sigma$ changes between rows. The
reported rule is not the best entry in every column: $2$\,s is $0.41$ points higher on
Charades train but $1.51$ lower on ActivityNet Qwen3-VL, and no fixed
bandwidth is best across all three ActivityNet backbones. For the $2$- and
$3$-second alternatives, the largest deviation from the reported rule is $1.28$
on Charades and $1.51$ on ActivityNet. Removing smoothing costs $5.45$ points
on Charades train and up to $4.15$ across the reported test backbones. Values
are recomputed from six-decimal saved scores. The $D/15$ row matches every
Charades test result to $0.00$. It differs from the original Charades-train and
ActivityNet evaluations by at most $0.17$ because rounding creates rare ties
on flat curves. Percentages.}
\label{tab:sigma-single}
\end{table}

\paragraph{Multi-interval aggregation sweep.}
Table~\ref{tab:sigma-sweep} separates the two length-dependent choices on
Qwen2.5-VL validation. Replacing the length-proportional smoothing with a fixed
clip-scale kernel lifts mAP from $20.76$ to $36.77$, and the per-curve Otsu
threshold lifts it to $42.46$. The gain concentrates in the mid-length band that
the global mean had over-extended, while short moments stay near the floor. This
validation-selected aggregation is then frozen for the test results.

\begin{table}[t]
\centering
{\renewcommand{\arraystretch}{1.1}\small
\setlength{\tabcolsep}{6pt}
\begin{tabular}{lcccc}
\toprule
\textbf{Aggregation} & \textbf{mAP} & \textbf{short} & \textbf{mid} & \textbf{long} \\
\midrule
mean threshold, $\sigma{=}\lfloor D\rfloor/15$ & 20.76 & 0.27 & 11.88 & 43.05 \\
mean threshold, $\sigma{=}3$s & 36.77 & 3.21 & 38.51 & 46.36 \\
mean threshold, $\sigma{=}2$s & 39.74 & 5.05 & 44.39 & 44.92 \\
Otsu threshold, $\sigma{=}3$s \textit{(selected)} & \textbf{42.46} & 5.39 & \textbf{47.82} & \textbf{46.50} \\
\bottomrule
\end{tabular}}
\caption{Aggregation selection on QVHighlights \texttt{validation} with Qwen2.5-VL-7B
(average mAP and duration-stratified mAP). Every row uses the same run
filtering and ranking procedure, so the rows differ only in the smoothing
bandwidth and the threshold. The length-proportional smoothing
$\sigma{=}\lfloor D\rfloor/15$ inflates to roughly $10$ seconds on a $150$-second video and
erases the moment; a fixed clip-scale kernel
($\sigma{=}3$s, matched to the clip) removes most of the loss ($+16.0$), and
replacing the per-video mean with a per-curve Otsu threshold at the same
bandwidth adds $+5.69$. The gain concentrates in the mid-length band that the
global mean was over-extending; moments shorter than a clip stay near the
floor. $\sigma{=}2$s is reported as a sensitivity check. The selected aggregation
(Otsu, $\sigma{=}3$s) is frozen unchanged for the test results of
Table~2 of the main paper.}
\label{tab:sigma-sweep}
\end{table}

\paragraph{Verifier prompt wording.}
Table~\ref{tab:prompt} repeats the pipeline on a fixed random 800-query validation
subset with four verifier instructions, holding the clips and the frozen aggregation
identical. The strict wording was written before this comparison and was not selected
from it. The strict default-\texttt{No} wording is best at $43.56$ mAP; a neutral
symmetric wording costs $2.2$, a minimal one $3.8$, and a paraphrase of the strict
intent $4.4$. No variant collapses, and the spread is comparable to switching
between aggregation variants in Table~\ref{tab:sigma-sweep}, so the grounding signal
comes from the verification task itself. The asymmetric instruction contributes a
consistent but second-order sharpening of the score separation.

\begin{table}[t]
\centering
{\renewcommand{\arraystretch}{1.1}\small
\setlength{\tabcolsep}{5pt}
\begin{tabular}{lccc}
\toprule
\textbf{Verifier prompt} & \textbf{MR mAP} & \textbf{R1@0.5} & \textbf{R1@0.7} \\
\midrule
strict default-\texttt{No} \textit{(used)} & \textbf{43.56} & \textbf{70.12} & \textbf{52.88} \\
neutral (``is this present?'')             & 41.35 & 68.00 & 50.25 \\
short (``does the video show'')            & 39.78 & 68.38 & 48.25 \\
paraphrase of the strict intent            & 39.19 & 67.12 & 47.50 \\
\bottomrule
\end{tabular}}
\caption{Verifier-prompt wording on an 800-query QVHighlights validation subset
(Qwen2.5-VL, frozen aggregation). Every wording stays within $4.4$ mAP of the one used,
so the method does not depend on one crafted sentence; the ordering consistently
favours the strict default-\texttt{No} instruction, which widens the score
separation between in-moment and out-of-moment clips.}
\label{tab:prompt}
\end{table}

\begin{figure}[t]
\centering
\includegraphics[width=\columnwidth]{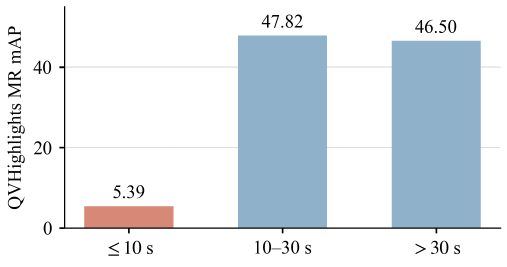}
\caption{QVHighlights validation moment-retrieval mAP by ground-truth moment
duration using Qwen2.5-VL-7B and the reported REZE configuration. Percentages.}
\label{fig:duration-breakdown}
\end{figure}

\paragraph{Short moments are detected, not delimited.}
Figure~\ref{fig:duration-breakdown} groups QVHighlights validation moments by
duration. Moment-retrieval mAP is $5.39$ for moments no longer than $10$
seconds, compared with $47.82$ and $46.50$ for the two longer groups.
The near-floor moment-retrieval mAP on short moments shows that localisation fails on
short moments; on its own it does not say whether recognition fails too. A direct
check on the native per-clip scores separates the two. On the $45$ test queries
whose ground-truth windows are all shorter than $10$ seconds, clips overlapping
the ground truth outscore the remaining clips of the same video with a per-query
AUROC of $0.84$--$0.87$ on all four backbones (per-query median GT-to-non-GT score
ratio $2.2\times$--$22.9\times$). Restricting to the strictest case, the $326$
individual ground-truth windows no longer than $2$ seconds, shorter than the
$3$-second clip itself, still gives a per-window AUROC of $0.71$--$0.76$, with
$72$--$82\%$ of windows scored above their video's non-ground-truth clips. The
verifier therefore detects most sub-clip moments; what a single clip-level
score cannot supply is their sub-clip boundary, which is the resolution argument
of Section~4.3 in the main paper. This post hoc diagnostic uses the reported
test predictions and does not affect model or parameter selection.

\paragraph{Clip length.}
Table~\ref{tab:seglen} varies the clip length $\tau \in \{2,3,4,5\}$\,s on the
validation split, checking afterwards how sensitive the results are to the
$\tau{=}3$\,s fixed in advance for the main paper, keeping six frames per clip so the frame rate
is $6/\tau$. Under
the frozen $\sigma{=}3$s Otsu aggregation, moment-retrieval mAP falls steadily from
$43.17$ at $\tau{=}2$s to $40.53$ at $\tau{=}5$s, and the loss is concentrated in the
short moments while the mid and long ranges stay roughly flat. This is the
clip-length resolution limit of Section~4.3 in the main paper seen from the clip side:
a coarser clip cannot place a short boundary. The finest clip length $\tau{=}2$s is
marginally better than $\tau{=}3$s (under $0.7$ mAP) but issues $1.5\times$ the
verification calls, so $\tau{=}3$s is a resolution--compute compromise rather than a
tuned optimum. Matching the smoothing to the clip ($\sigma{=}\tau$) rather than fixing
it at $3$s is no better, and worse at coarse clip lengths, because a wide clip with wide
smoothing over-blurs short moments: their mAP falls from $9.31$ at
$\tau{=}2$s to $1.80$ at $\tau{=}5$s.

\begin{table}[t]
\centering
{\renewcommand{\arraystretch}{1.1}\small
\setlength{\tabcolsep}{6pt}
\begin{tabular}{lccccc}
\toprule
\textbf{Clip $\tau$} & \textbf{fps} & \textbf{MR mAP} & \textbf{short} & \textbf{mid} & \textbf{long} \\
\midrule
2\,s                    & 3.0 & 43.17 & 6.06 & 49.08 & 46.63 \\
3\,s \textit{(used)}     & 2.0 & 42.46 & 5.39 & 47.82 & 46.50 \\
4\,s                    & 1.5 & 40.80 & 4.37 & 46.50 & 44.69 \\
5\,s                    & 1.2 & 40.53 & 4.39 & 45.60 & 45.13 \\
\bottomrule
\end{tabular}}
\caption{Clip-length sweep on QVHighlights \emph{validation} with
Qwen2.5-VL-7B. Each clip keeps six frames, so the frame rate is $6/\tau$. Under the
frozen $\sigma{=}3$s Otsu aggregation, moment-retrieval mAP degrades steadily as the
clip lengthens, and the loss concentrates in short moments while mid and long
stay roughly flat: a coarser clip cannot place a short boundary, the same resolution
limit seen in Section~4.3 of the main paper. The finest clip length $\tau{=}2$s is only
$0.7$ mAP better than $\tau{=}3$s yet issues $1.5\times$ the verification calls, so
$\tau{=}3$s is a resolution--compute compromise rather than a tuned peak. Scaling the
smoothing to the clip ($\sigma{=}\tau$) instead of fixing it gives $42.60/42.46/39.87/38.17$
mAP for $\tau{=}2/3/4/5$, no better and worse at the coarse end.}
\label{tab:seglen}
\end{table}

\subsection{D. Direct-Generation Diagnostics}
\label{app:direct}

Each Direct response is sorted into one of five groups, and the
\miou{} over the cleanly parsed (``reasonable'') group is reported separately
(Table~\ref{tab:failure-mode}): (i)~parse fail, (ii)~short interval ($<0.3$\,s),
(iii)~out-of-bound, (iv)~collapsed start (parsed interval at exactly $0.0$, a
template-collapse heuristic; legitimate early predictions are scored normally in
the main \miou{}), (v)~reasonable.

\begin{table*}[t]
\centering
{\small
\begin{tabular}{lccccccc}
\toprule
\textbf{Model} & \textbf{Direct \miou} & \textbf{Parse Fail} & \textbf{Short Interval} & \textbf{OOB} & \textbf{Collapsed Start} & \textbf{Reasonable} & \textbf{Reasonable \miou} \\
\midrule
Qwen2-VL-7B    & 6.48 & 0.7\%  & 0.0\% & \textbf{52.7\%} & 28.1\% & 18.5\% & 24.88 \\
Qwen2.5-VL-7B  & 42.74 & 0.0\%  & 0.4\% & 11.9\%          & 2.8\%  & 84.9\% & 43.50 \\
Qwen3-VL-8B    & 53.30 & 0.1\%  & 0.0\% & 1.0\%           & 11.0\% & \textbf{87.9\%} & \textbf{54.48} \\
InternVL3.5-8B & 30.17 & 0.0\%  & 0.4\% & 5.8\%           & 0.4\%  & 93.4\% & 31.47 \\
\bottomrule
\end{tabular}}
\caption{Where direct generation goes wrong on Charades-STA. Each row sorts the
3720 Direct responses into five groups. ``Reasonable'' means the response
parsed correctly and fell inside the video duration; the other four are
failures. ``Reasonable \miou{}'' is the \miou{} over the reasonable group
alone, that is, how accurate the prediction is when the model does get the
format right. All percentages of N=3720.}
\label{tab:failure-mode}
\end{table*}

\paragraph{Interface failures account for a large share of Direct errors.}
Although the Direct prompt explicitly requests seconds, Qwen2-VL-7B's $6.48$
Direct \miou{} suffers a $52.7\%$ out-of-bound rate because it emits frame
indices (``From Frame 400 to Frame 599''); restricting to the
reasonable group raises its \miou{} to $24.88$, an 18-point jump. The reasonable
subset is a conditional sample, and $24.88$ still sits far below REZE's $51.49$,
so on Qwen2-VL the interface accounts for a large share of the error rather than
all of it. The strongest support for the interface reading comes from the newer
models: Qwen2.5-VL and Qwen3-VL close the parse/OOB gap ($0.0\%$ parse fail;
$11.9\%$ and $1.0\%$ OOB), and their reasonable \miou{} ($43.50$, $54.48$) approaches REZE
($54.22$, $55.66$), within $1.18$ for Qwen3-VL.
InternVL3.5-8B shows subtle template collapse: a $93.4\%$ reasonable rate yet
$30.17$ \miou{}, because many of its parsed responses fall into a few fixed
``$K$ to $K{+}5$ seconds'' templates. Across all four models, Direct requires the
backbone to verbalise absolute time in a parseable, query-conditional form; REZE removes that
requirement and applies unchanged across their different native video and
timestamp interfaces.

\begin{figure*}[t]
\centering
\includegraphics[width=\textwidth]{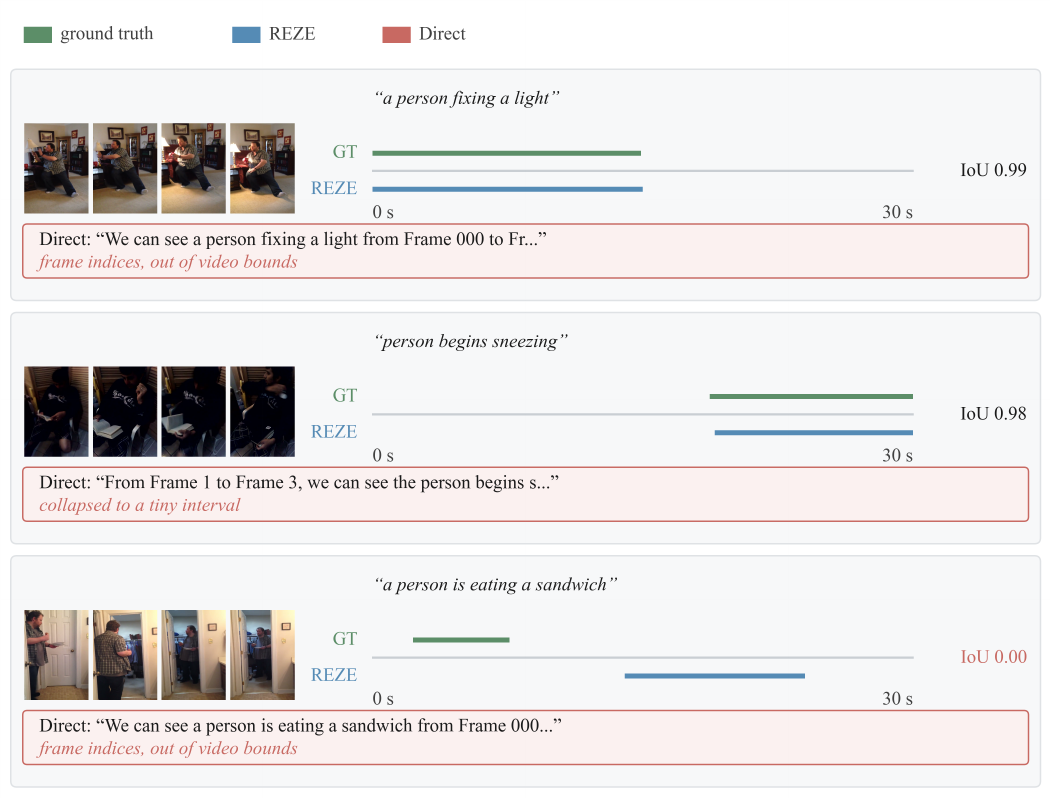}
\caption{Qualitative comparison on Charades-STA with Qwen2-VL-7B. Each row shows a
frame strip, the query, and the ground-truth (green) and \reze{} (blue) intervals
with IoU. Direct emits frame indices or narration rather than parseable seconds and
cannot be placed on the timeline (red diagnostic box). The first two rows are successes where
\reze{} recovers the interval while Direct fails the output interface; the last is
an honest \reze{} failure on a short early action.}
\label{fig:qualitative}
\end{figure*}

\subsection{E. Efficiency}
\label{app:efficiency}

REZE issues $\lceil D/\tau \rceil$ verification calls per query, about eleven for a
$30$-second Charades video and fifty-one for a $150$-second QVHighlights video,
versus one generation call for Direct. Table~\ref{tab:cost} reports the measured
cost of both modes on Qwen2.5-VL. Throughput is the process-level wall-clock of a
fixed random 200-query subset run through the actual vLLM pipeline (engine
initialisation amortised over the subset), with each mode at its stable
concurrency: 32 concurrent sequences for REZE and 8 sequences with request batches
of 4 for Direct, both at 2\,fps and 0.85 GPU-memory fraction. Peak memory comes
from a separate single-stream HuggingFace measurement; the reported figure is the
transient increment above the resident 16.6\,GB of loaded model weights, taken with
a fresh baseline before every call.

The headline trade-off is summarised in Table~1C of the main paper and measured
in full in Table~\ref{tab:cost}: REZE processes $2.2$--$2.3\times$ more input tokens per
query, its batched throughput is $1.4$--$1.5\times$ lower, and in exchange every
call has a bounded footprint that does not depend on the video length. Two
further observations follow. First, the per-call cost is constant in a strong sense:
clip-level throughput stays at $2.6$--$2.8$ calls per second on both benchmarks,
so the per-query cost scales only through the call count $\lceil D/\tau\rceil$.
Second, without batching the estimated serial latency of REZE is $5.4$\,s per
Charades query and $27.5$\,s per QVHighlights query, against a batched effective
$3.8$\,s and $19.1$\,s; these are medians over 20 real queries, with decoding,
preprocessing, and generation timed per clip and multiplied by the call count.
One caveat applies to the wall-clock comparison. The REZE pipeline decodes each
clip separately with OpenCV, about a quarter of the serial estimate, whereas the
Direct pipeline
decodes each video once with decord, so the gap in pure model compute is
smaller than the wall-clock gap suggests.

\begin{table*}[t]
\centering
{\renewcommand{\arraystretch}{1.1}\small
\setlength{\tabcolsep}{4pt}
\begin{tabular}{lcccc}
\toprule
\textbf{Setting} & \textbf{calls} & \textbf{input tokens} & \textbf{queries/s} & \textbf{peak $\Delta$VRAM/call} \\
\midrule
Direct, Charades   & 1          & $\sim$5.6k          & 0.41  & 0.5--1.7\,GB \\
REZE, Charades     & $\sim$11   & 1.1k\,$\times$\,11  & 0.27  & 0.26\,GB \\
Direct, QVH        & 1          & $\sim$31.5k         & 0.073 & 6.6--7.2\,GB \\
REZE, QVH          & $\sim$51   & 1.4k\,$\times$\,51  & 0.052 & 0.31\,GB \\
\bottomrule
\end{tabular}}
\caption{Measured per-query inference cost, all on Qwen2.5-VL-7B. Throughput is
the process-level wall-clock of a 200-query subset through the vLLM pipeline.
Peak $\Delta$VRAM is the transient peak of a single active request above the
$16.6$\,GB of loaded weights (range over sampled queries), not a sum across calls:
sequential REZE calls reuse the same footprint. Direct's tokens and memory grow
roughly linearly with video length, whereas each REZE call is constant and only
the call count grows.}
\label{tab:cost}
\end{table*}

\subsection{F. Qualitative Results}
\label{app:qual}

\begin{figure*}[t]
\centering
\includegraphics[width=\textwidth]{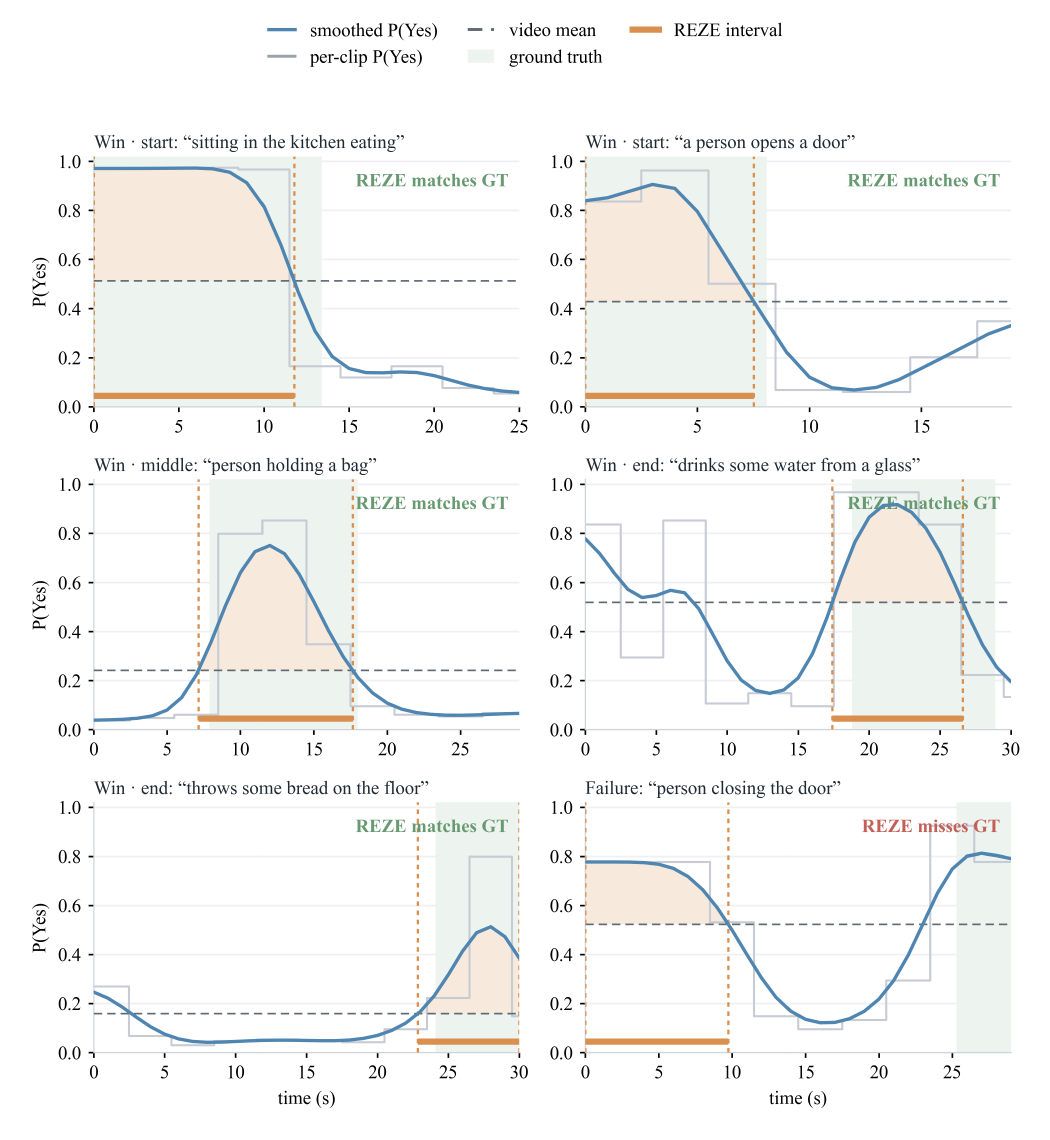}
\caption{Real $P(\text{Yes})$ curves on Charades-STA (Qwen2-VL-7B) for six queries
whose ground truth falls at the start, middle, and end of the video, plus one
failure. Grey steps are per-clip $P(\text{Yes})$; the blue line is the smoothed
curve; the dashed line is the video-mean threshold. Ground truth is shaded green
and the \reze{} interval (the mean-shifted maximum-subarray interval of
Section~3.2) is shaded orange, read from the same smoothed curve; the dashed
video-mean line is the zero level of the mean-shifted objective.}
\label{fig:curves-appendix}
\end{figure*}

Figure~\ref{fig:curves-appendix} shows six real per-clip $P(\text{Yes})$ curves on
Charades-STA (Qwen2-VL-7B), with the ground truth at the start, middle, and end of
the video plus a failure. In the five successes the aggregation tracks the action; in
the failure (\texttt{0QAZ7\_1}) \reze{} selects the wrong run, far from the short
ground-truth segment at the very end.

\subsection{G. Limitations and Outlook}
\label{app:limits}

REZE has several limitations that warrant future investigation:

\begin{itemize}
    \item \textbf{Computational Trade-offs:} REZE requires approximately $\lceil D/\tau\rceil$ forward passes per query, resulting in a $1.4$--$1.5\times$ lower batched throughput compared to direct generation (Appendix~E). In return, its memory footprint remains constant regardless of video length.

    \item \textbf{Cross-clip Relations:} The aggregation preserves the temporal order of the clip scores, but the verifier scores each clip independently. It therefore cannot model cross-clip relations required for action ordering, counting, or motion direction.

    \item \textbf{Dataset and Backbone Edge Cases:} Our ActivityNet evaluation is limited to the Qwen family, relying on Charades-STA and QVHighlights for cross-backbone evidence. On single-interval long videos within these datasets, Qwen3-VL direct generation marginally outperforms REZE ($36.76$ vs.\ $35.43$ \miou{}). This is not caused by the aggregation: applying a clip-scale aggregation changes REZE by at most $1.2$ \miou{} and fails to close the gap (Appendix~B).

\end{itemize}

\end{document}